\documentclass[10pt,twocolumn,letterpaper]{article}

\usepackage{3dv}
\usepackage{times}
\usepackage{epsfig}
\usepackage{graphicx}
\usepackage[subpreambles]{standalone}
\usepackage{amsmath}
\usepackage{amssymb}
\graphicspath{{images/}}
\DeclareGraphicsExtensions{.pdf,.jpg,.png}
\usepackage[caption=false,font=footnotesize]{subfig}

\usepackage{ptkabbr}
\usepackage{ptkref}
\usepackage{ptkmath}
\usepackage{ptkcolor}

\newcommand{\RGBD}{\mbox{RGB-D}\xspace}

\newcommand{\OurApproach}{\mbox{HDR-SLAM}\xspace}

\newcommand{\Asus}{Asus Xtion Live Pro\xspace}

\usepackage[pagebackref=true,breaklinks=true,letterpaper=true,colorlinks,bookmarks=false]{hyperref}
\usepackage[capitalize,noabbrev]{cleveref}

\threedvfinalcopy

\def\threedvPaperID{176}

\ifthreedvfinal\fi

\begin{document}
\title{
High Dynamic Range SLAM with Map-Aware Exposure Time Control
}

\author{Sergey V. Alexandrov, Johann Prankl, Michael Zillich and Markus Vincze \\
        Vision4Robotics Group, ACIN, TU Wien \\
        {\tt\small \{alexandrov, prankl, zillich, vincze\}@acin.tuwien.ac.at}
}

\maketitle
\thispagestyle{empty}

\begin{abstract}
The research in dense online 3D mapping is mostly focused on the geometrical accuracy and spatial extent of the reconstructions. Their color appearance is often neglected, leading to inconsistent colors and noticeable artifacts. We rectify this by extending a state-of-the-art SLAM system to accumulate colors in HDR space. We replace the simplistic pixel intensity averaging scheme with HDR color fusion rules tailored to the incremental nature of SLAM and a noise model suitable for off-the-shelf RGB-D cameras. Our main contribution is a map-aware exposure time controller. It makes decisions based on the global state of the map and predicted camera motion, attempting to maximize the information gain of each observation. We report a set of experiments demonstrating the improved texture quality and advantages of using the custom controller that is tightly integrated in the mapping loop.
\end{abstract}

\section{Introduction}
\label{sec:intro}

\begin{figure}[!t]
  \centering
  \subfloat{\includegraphics[trim={100 335 110 290},clip,width=\columnwidth]{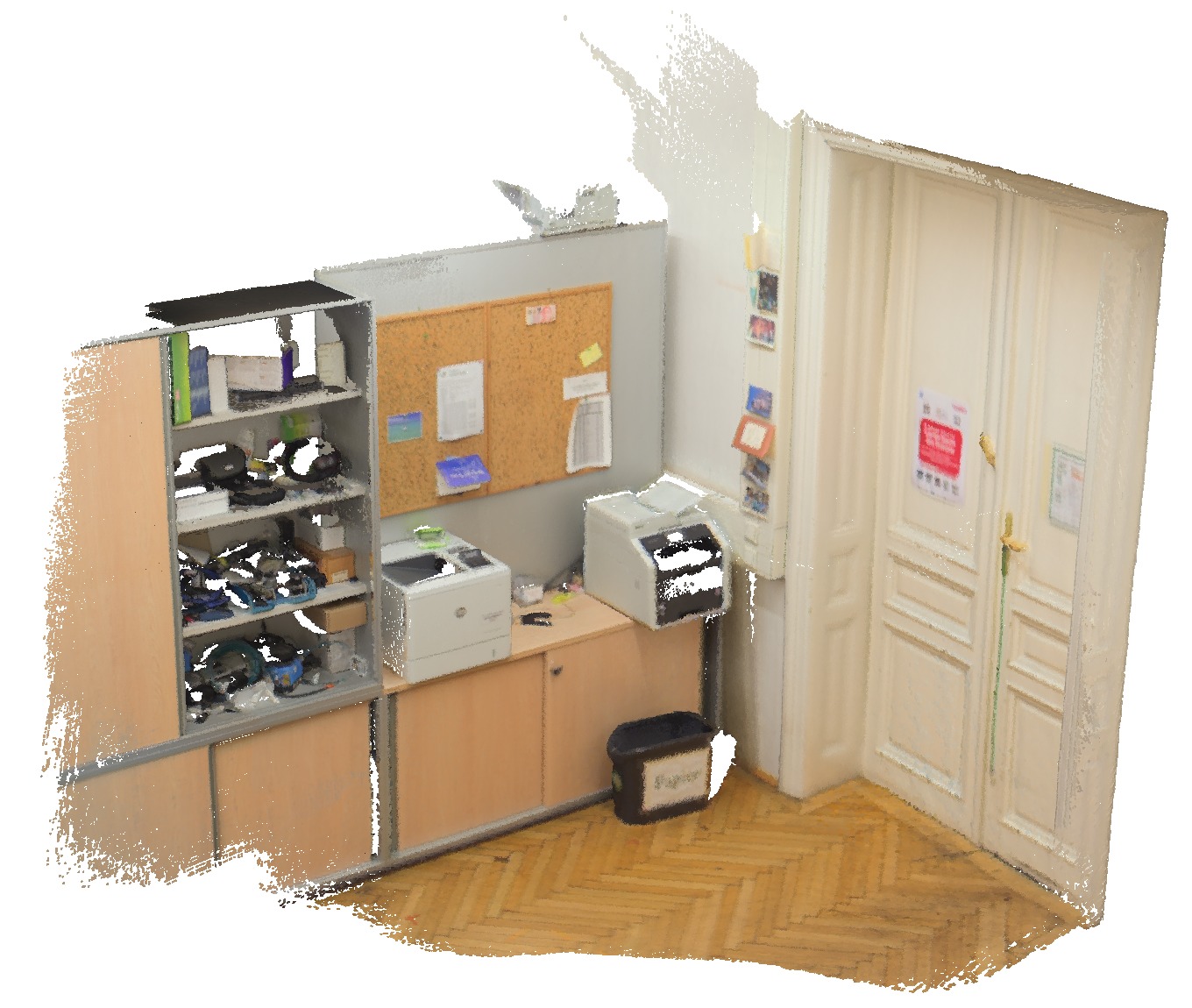}%
  \label{fig:cmp_entrance_hdr}}
  \vspace*{-6pt}
  \subfloat{\includegraphics[trim={100 335 110 290},clip,width=\columnwidth]{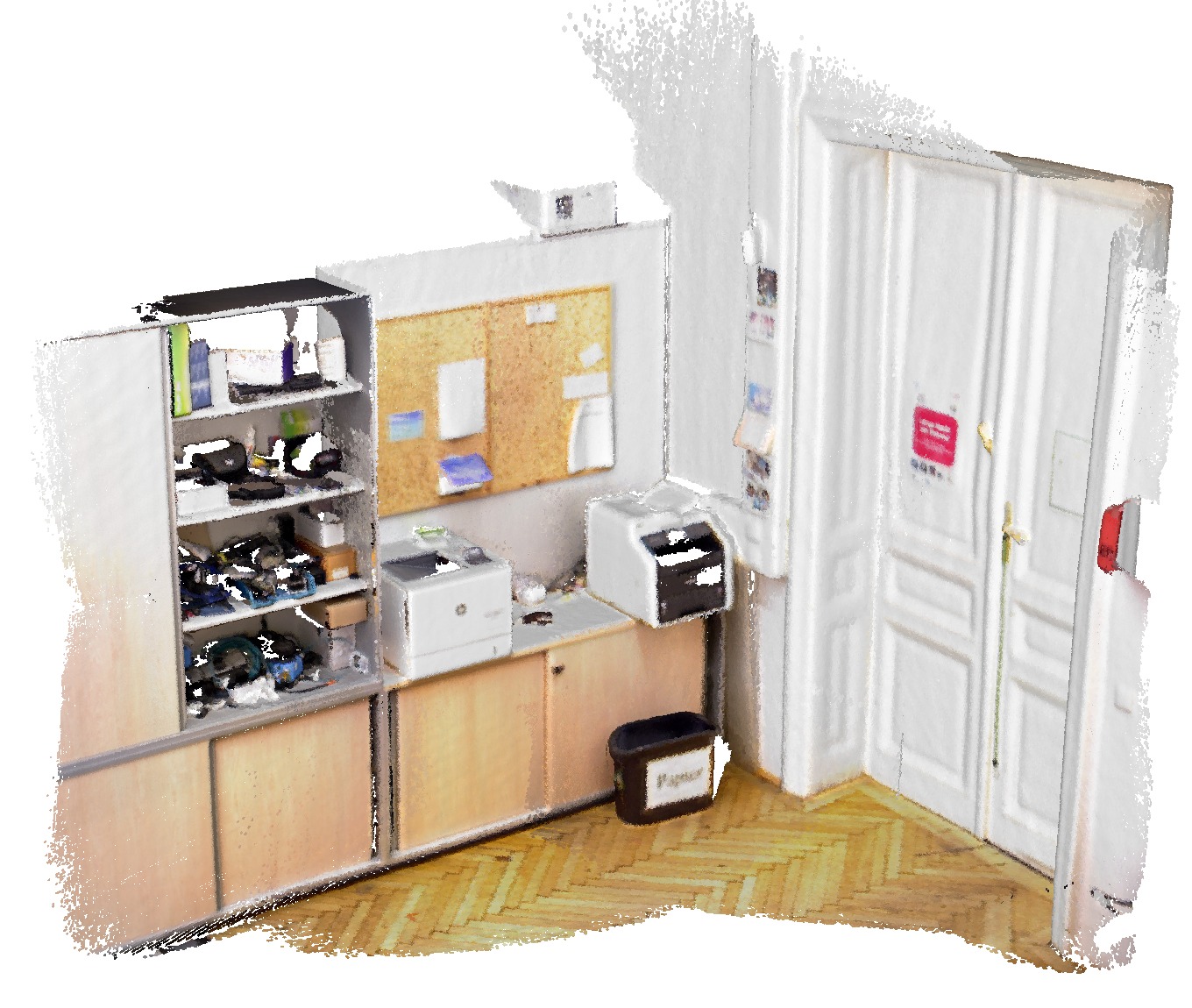}%
  \label{fig:cmp_entrance_ldr_fix}}
  \caption{
    Top: an office scene reconstructed in HDR with the proposed exposure time controller. Note the difference in color between the wooden panel on the left, the gray wall, and the milky white door on the right. Bottom: the same scene reconstructed in LDR with ElasticFusion using a fixed exposure time setting. The dynamic range of the scene exceeds that of an LDR image; as a result the panel, wall, and door were overexposed and appear to have the same color in the reconstruction.
  }
  \label{fig:cmp-entrance}
\end{figure}

The introduction of \RGBD cameras to the consumer market spawned a revolution in 3D vision. The progress in camera tracking and online dense reconstruction reached impressive milestones in terms of tracking accuracy and robustness \cite{Zhou2015,Hsiao2017}, as well as in scale and geometric fidelity of large reconstructions \cite{Dai2017,Whelan2015}.

The quality of color in dense 3D mapping received less attention despite being equally important in applications such as virtual and augmented reality, semantic analysis, and recognition. Recent works demonstrated an improvement in the texture quality through radiometric calibration of the camera \cite{Alexandrov2016}, subsequent offline texture map generation and optimization of camera parameters \cite{Zhou2014,Jeon2016}, online estimation of surface reflectance \cite{Kim2017} and light sources within the reconstructed scenes \cite{Whelan2016}.

As the 3D reconstructions grow in scale, the problem of low dynamic range (LDR) of camera sensors is becoming more evident. The 24-bit colors are inadequate for the representation of the wide range of light intensities found in large real-world scenes. Thus a promising yet relatively unexplored avenue is the application of High Dynamic Range (HDR) imaging techniques in 3D reconstruction \cite{Meilland2013}. This will enable the acquisition of radiance maps of scenes with a dynamic range greatly exceeding LDR cameras, achieving more visually appealing renderings of the reconstructed models as well as more consistent textures (see \fref{cmp-entrance}).

HDR imaging is a well-understood topic in photography. Its basic form involves taking multiple LDR images at different exposure times and combining them into a composite radiance map. The problems addressed in the HDR imaging literature include calibration of the camera response function \cite{Debevec1997}, noise modeling and sample weighting schemes \cite{Granados2010}, exposure time selection \cite{Hasinoff2010}, coping with inconsistent measurements due to camera jitter or scene dynamics \cite{Zimmer2011}.

The application of HDR imaging techniques in online 3D reconstruction is not straightforward. The domain is not restricted by the image pixel grid and grows arbitrarily in space. The limited memory and processing time budgets permit only a small pre-allocated space per map point, excluding the possibility to store multiple observations. Together with the need for online feedback, this dictates the usage of incremental estimation schemes instead of batch optimization. Furthermore, in HDR imaging camera motion is regarded as an unwanted effect, whereas in SLAM the moving camera is the \emph{modus operandi}, leading to an increased amount of outlier measurements. It also adds a temporal aspect to the problem, requiring efficient control of exposure time to ensure that the color of the map points is estimated with sufficient accuracy before they leave the field of view of the camera.

To overcome these problems, we propose \OurApproach \footnote{Our system extends ElasticFusion \cite{Whelan2015} and is hosted as a GitHub fork \url{https://github.com/taketwo/ElasticFusion/tree/hdr}}, an online incremental 3D reconstruction approach that captures the scene appearance in HDR colors. We contribute:

\begin{itemize}
  \item a map-aware exposure time controller integrated in the SLAM loop aiming to maximize the information gain of each
observation;
  \item HDR color fusion rules tailored to the incremental nature of SLAM reconstruction;
  \item a noise model for off-the-shelf \RGBD cameras.
\end{itemize}

We evaluate the performance of \OurApproach with a static camera, where a comparison with batch-processed ``ground truth'' reconstruction can be made. The quantitative results show that the proposed controller outperforms other baselines. Furthermore, we demonstrate side-by-side comparisons of the HDR reconstructions and those obtained with a baseline LDR SLAM approach, showing significantly improved texture quality.

The rest of the paper is structured as follows. \cref{sec:pre} gives the necessary background in HDR imaging and \cref{sec:rw} covers the related work. \cref{sec:arch} presents the system architecture, followed by the description of camera noise model (\cref{sec:model}), incremental HDR reconstruction (\cref{sec:inc}), exposure time controller (\cref{sec:ctrl}), and additional implementation details (\cref{sec:impl}). The system is evaluated in \cref{sec:eval} and the paper is concluded in \cref{sec:ciao}.

\section{HDR imaging foundations}
\label{sec:pre}

Application of HDR color acquisition techniques in the context of dense 3D reconstruction requires understanding of the image formation process, its associated noise sources and limitations, as well as the core ideas behind HDR imaging. This section serves as a brief introduction; an in-depth discussion can be found in \cite{Sen2016,Kolb1995,Healey1994}.

\subsection{Image formation process}

Scene surfaces emit or reflect light rays; the amount of radiant power that they carry (\emph{radiance}) determines the appearance of the scene to the observer. To capture the appearance, a camera maps radiances into image pixel \emph{intensities} through a sequence of nonlinear transformations.

First, the light rays enter the camera aperture, pass through the lens system, and land on a lattice of photon wells. The radiant power density (\emph{irradiance}) incident on the well surface is integrated over the time $t$. The accumulated radiant energy (\emph{exposure}) is then converted into a voltage, amplified, digitized, and mapped through a nonlinear radiometric response function into a pixel intensity.

The output value at an image location $\mathbf{u}$ is given by
\begin{equation}
  Z(\mathbf{u}) = f(tE(\mathbf{u})),
  \label{eqn:rif}
\end{equation}
where $E$ are pixel irradiances and $f : \mathbb{R}\rightarrow\left\{0,\ldots,255\right\}$ is a composition of the
amplification, digitization, radiometric response, and quantization, which will be further referred to as the camera response function (CRF).

Radiance and irradiance are directly proportional, the constant of proportionality being dependent on the properties of the lens system. In most applications the absolute scale of radiance is not important and these terms are used interchangeably. However, due to several factors, collectively referred to as vignetting effects \cite{Goldman2005}, the coefficient of proportionality is not uniform across the image plane, typically exposing radial fall-off from the center to the edges. This is important in the context of our work, thus we distinguish between the two terms and define
\begin{equation}
  E(\mathbf{u}) = L(\mathbf{u})V(\mathbf{u}),
  \label{eqn:irradiance}
\end{equation}
where $L$ are pixel radiances and $V$ are image location dependent vignetting attenuation coefficients.

\subsection{Noise sources}

The image formation process is affected by multiple error sources~\cite{Healey1994}. Due to the quantum nature of light, the number of photo-induced electrons collected at a photon well follows a Poisson distribution; its uncertainty is called photon shot noise (PSN). Dark current contributes thermo-induced electrons that add up to the accumulated energy. Several other noise sources associated with conversion from charge to digital values are collectively referred to as readout noise. Due to imprecisions in the manufacturing process, for different pixels the photo-response is not uniform (PRNU) and so is the amount of dark current (DCNU).

\subsection{Dynamic range}

The noise and the finite capacity of photon wells limit the range of radiant energies $\left[x^{\text{min}},x^{\text{max}}\right]$ that can be accumulated and detected by a camera. For any given exposure time $t$, this determines the \emph{detectable range} of irradiances
\begin{equation}
  \epsilon_{t} =
    \left[\frac{x^{\text{min}}}{t},\frac{x^{\text{max}}}{t}\right] = 
    \left[e^{\text{min}}_t,e^{\text{max}}_t\right].
  \label{eqn:detectable-range}
\end{equation}
Scene points inducing irradiance below or above will appear as black (underexposed) or white (overexposed) pixels respectively. The ratio of the boundary values is independent of the exposure time and is called the \emph{dynamic range}.

Let $\mathcal{T}$ be the set of all supported exposure time settings. The union of their corresponding detectable ranges
\begin{equation}
  \epsilon_{\text{sys}} = 
  \bigcup_{t\in\mathcal{T}}\epsilon_{t} = 
  \left[\frac{x^{\text{min}}}{\left\lceil \mathcal{T}\right\rceil },\frac{x^{\text{max}}}{\left\lfloor \mathcal{T}\right\rfloor }\right] =
  \left[e^{\text{min}},e^{\text{max}}\right]
  \label{eqn:system-detectable-range}
\end{equation}
is the \emph{effective detectable range} of the camera. Irradiances outside of this range can not be measured.

\subsection{HDR imaging}

The goal of HDR imaging is to recover an irradiance image of a scene in its full dynamic range. A set of LDR images $Z_i$ is taken at different exposure times $t_i$. Each of them is converted into irradiance image $\hat{E}_i$ according to \eref{rif}. The irradiance estimates in these images are in a linear space at a fixed common scale, which allows us to combine them using a weighted average:
\begin{equation}
  \bar{E}
    =
    \frac{\sum_{i=1}^{n}W_i\hat{E}_{i}}{\sum_{i=1}^{n}W_i},
  \label{eqn:hdr}
\end{equation}
where $n$ is the number of images and $W_i$ are per-pixel confidence weights. The purpose of the weights is to discard poorly exposed pixels that carry no information and to emphasize more reliable samples.

Various weighting schemes were proposed in the literature. In the early work somewhat ad-hoc options were used, including the gradient of CRF \cite{Mann1995} and a hat function \cite{Debevec1997}. Later, Kirk and Andersen~\cite{Kirk2006} characterized several other weighting schemes, concluding that the variance-based weighting gives best lower bound on signal-to-noise ratio. Granados \etal[Granados2010] presented a rigorous camera noise model that takes into account both temporal and spatial sources. They note that variance-based weighting indeed yields Maximum Likelihood Estimate (MLE), however due to the photon shot noise, the variance of a sample depends on the true irradiance. This introduces a circular dependency that they propose to solve with iterative estimation.

\section{Related work}
\label{sec:rw}

\subsection{HDR-aware mapping}

The idea of using HDR colors in the context of 3D reconstruction with \RGBD cameras was pioneered by Meilland \etal[Meilland2013]. In their visual SLAM system the scene is modeled by a graph of super-resolved keyframes. Each keyframe is a product of HDR-aware fusion of a sequence of aligned camera frames. They model the camera response with the gamma function and ignore the vignetting effects. Unlike the classical HDR imaging where exposure time of each frame is preselected and known, they rely on the built-in auto exposure controller (AEC) function. This means they need to estimate the relative exposure time change jointly with the camera transform during camera tracking. 

Recently, Li \etal[Li2016] extended a volumetric SLAM framework to accumulate colors in HDR space. They also use AEC, but solve the camera tracking problem in the normalized radiance space that is independent of exposure time. Once the frames are aligned, the exposure time change is computed as a weighted average of the radiance ratios between corresponding pixels. They use a calibrated CRF and variance-based weights to fuse new color measurements into the global volumetric representation, but do not account for the vignetting effects.

Zhang \etal[Zhang2016] noted that frame-by-frame estimation of exposure time changes suffers from drift accumulation. They propose an offline method where per-frame exposure times and point radiances are the unknowns in a nonlinear optimization problem. Solving it allows to obtain globally optimal HDR textures for the reconstructed 3D model.

Unlike the mentioned works, we propose to actively control the exposure time. This means drift-free operation without the need for global optimization. Additionally, we use full radiometric calibration including vignetting effects.

\subsection{Exposure time control}

The problem of selecting a set of exposure times (\emph{bracketing set}) has been studied in the HDR imaging literature. Barakat \etal[Barakat2008] proposed several algorithms aimed to compute minimal bracketing sets. However, he was interested to obtain a single non-saturated observation per pixel and did not consider other properties of reconstruction such as signal-to-noise ratio (SNR). Hasinoff \etal[Hasinoff2010] investigated the problem of selecting exposure times and gains for noise-optimal HDR capture, however they assume that the distribution of radiances in the scene is known \emph{a priori}. Ilstrup and Manduchi \cite{Ilstrup2010} introduced an algorithm that determines the single optimal exposure time for a scene given one suboptimally exposed image. In the context of visual odometry, Zhang \etal[Zhang2017] proposed an active exposure controller that maximizes a gradient-based image quality metric. Differently from these works, our exposure time controller has no \emph{a priori} knowledge about the scene and aims to maximize the SNR of every reconstructed point.

\section{System architecture}
\label{sec:arch}

Our system builds upon a typical dense SLAM pipeline, where map updates are alternated with camera tracking \wrt the rendered view of the map. Specifically, we extended the open-source system of Whelan \etal[Whelan2015], where the map is represented by a surfel cloud. However, it should be noted that our approach can be combined with other map representations (\eg volumetric or keyframe-based).

A simplified system architecture diagram is given in \fref{architecture}. The components that are not essential for the topic of this paper (\eg loop detection, nonrigid map deformation) are not shown. The frames coming from the camera are radiometrically rectified using the camera model described in \cref{sec:model}. After the current pose of the camera is computed by the frame-to-model tracking module, the depth and rectified color images are fused into the map according to the rules outlined in \cref{sec:inc}. Next, the view of the map from the current pose is predicted, and this prediction is used by the exposure time controller as detailed in \cref{sec:ctrl}.

\begin{figure}[!t]
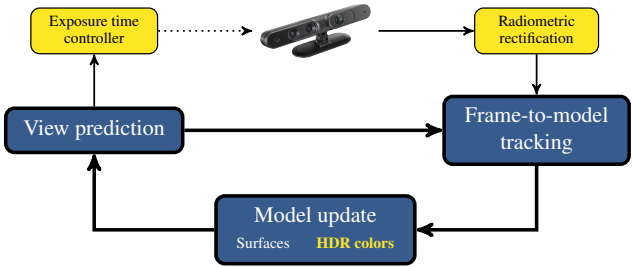

  \centering
  \includestandalone[width=\columnwidth]{architecture}
  \vspace*{1pt}
  \caption
  {
    Simplified system architecture diagram. The blue components are typical for dense SLAM systems and form the mapping loop (bold arrows). We introduce the yellow components to support acquisition of HDR colors. 
  }
  \label{fig:architecture}
\end{figure}

\section{Camera noise model}
\label{sec:model}

Online dense 3D reconstruction is dominated by \RGBD sensors. Typically, they are equipped with low-end color cameras that do not provide access to the raw measurements of radiant energy. Instead, a certain on-chip post-processing (\eg denoising, black frame subtraction) takes place. Consequently, accurate noise characterization as in \cite{Hasinoff2010,Granados2010} is not feasible. In the lack of knowledge about the camera internals, it was suggested to jointly model the noise sources with a compound Gaussian \cite{Robertson1999}. Liu \etal[Liu2008] proposed a model with two additive zero-mean components:
\begin{equation}
  Z(\mathbf{u}) = f(tE(\mathbf{u}) + n_s + n_c).
  \label{eqn:noise-simplified}
\end{equation}
The first component $n_s$ accounts for the noise dependent on the signal; its variance $\sigma^2_s$ is proportional to the exposure. The second component $n_c$ captures the independent noise sources and has a fixed variance. Our experiments with \Asus cameras (see \cref{ssec:ver}) have confirmed the general suitability of this model, however we found that the independent component can be neglected.

The model \eref{noise-simplified} ignores spatially varying error sources. However, it was demonstrated that vignetting effects are significant in \RGBD cameras; correcting them has a positive impact on both tracking accuracy and map quality \cite{Engel2016a,Alexandrov2016}. Therefore, we include vignetting effects and propose at the following model of individual pixel intensity:
\begin{equation}
  Z(\mathbf{u})=f(X(\mathbf{u})+n_{s}),
  \label{eqn:imaging-equation}
\end{equation}
where $ X(\mathbf{u})=tL(\mathbf{u})V(\mathbf{u})$ is the exposure, and its variance $\sigma_{X(\mathbf{u})}^{2}$ is proportional to $X(\mathbf{u})$ with some coefficient $a$. We derive an estimator for the radiance
\begin{equation}
  \hat{L}(\mathbf{u}) =
    \frac{f^{-1}(Z(\mathbf{u}))}{tV(\mathbf{u})} =
    \frac{X(\mathbf{u})}{tV(\mathbf{u})},
  \label{eqn:radiance}
\end{equation}
and its uncertainty
\begin{equation}
  \sigma_{L(\mathbf{u})}^{2} =
    \frac{\sigma_{X(\mathbf{u})}^{2}}{t^{2}V(\mathbf{u})^{2}} =
    \frac{at\hat{L}(\mathbf{u})V(\mathbf{u})}{t^{2}V(\mathbf{u})^{2}} =
    \frac{a\hat{L}(\mathbf{u})}{tV(\mathbf{u})}.
  \label{eqn:radiance-uncertainty}
\end{equation}

\subsection{Calibration}

Computation of pixel radiance using \eref{radiance} requires the function $g \equiv f^{-1}$ and the map of vignetting attenuation factors $V$ to be known. We use the method of Debevec \etal[Debevec1997] to obtain the former. As the absolute scale is not important, the calibrated response function is normalized such that $g(255) = 1$. The latter is calibrated using the method of Alexandrov \etal[Alexandrov2016]. It is worth noting that their method can not separate the vignetting effects from PRNU, jointly modeling them as single attenuation factor per pixel. \fref{calibration} shows the obtained radiometric calibration of one of the color channels of an \Asus camera.

\begin{figure}[!t]
  \centering
  \subfloat{\includegraphics[width=0.48\columnwidth]{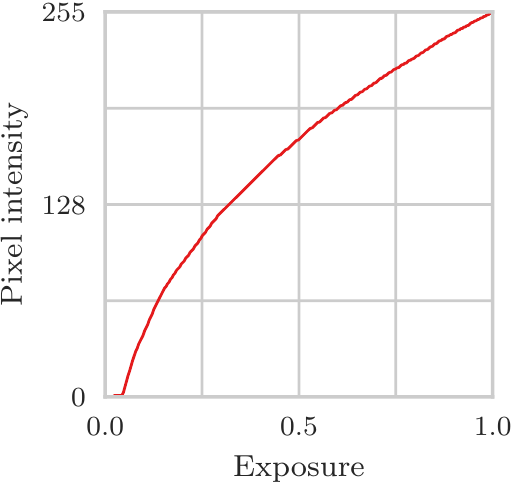}%
  \label{fig:rr}}
  \hfil
  \subfloat{\includegraphics[width=0.48\columnwidth]{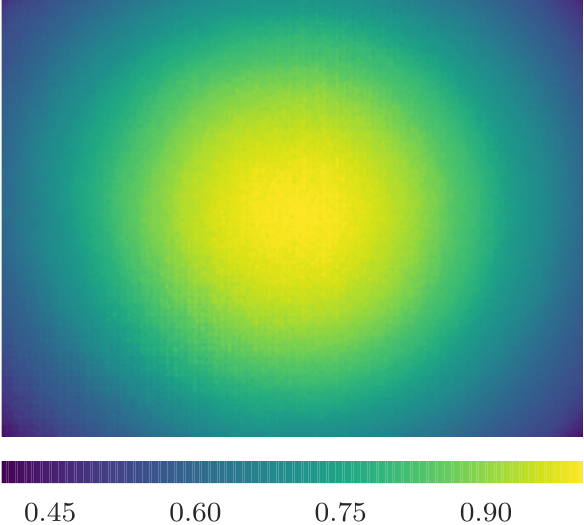}%
  \label{fig:vr}}
  \vspace*{5pt}
  \caption{
    Radiometric calibration of the red color channel of an \Asus camera. The camera response function (left) nonlinearly maps exposure to pixel intensity. The vignetting effects (right) introduce spatially varying attenuation; while the central pixels are not affected, the corners are up to two times darker.
  }
  \label{fig:calibration}
\end{figure}

\subsection{Verification}
\label{ssec:ver}

We verify the proposed noise model \eref{imaging-equation} by demonstrating that the noise variance is indeed linearly dependent on the signal. To show this, we fix the camera in front of a high dynamic range scene and select a bracketing set such that exposures span the whole effective detectable range of the camera. For every exposure time setting we capture 900 frames and compute the mean and variance of every pixel's exposure $g(Z(\mathbf{u}))$. Next we group these observations based on the mean exposure into bins spanning $[0..1]$ range. In each bin we select the median variance observation.

\fref{exposure-variance} shows the obtained relation between the exposure and variance. We observe a linear dependency which supports the signal-dependent noise component of our model. Furthermore, there is no noticeable offset along the y-axis, suggesting that the signal-independent component can be ignored. The drop in the variance near the maximum exposure is explained by the fact that the distribution is truncated due to the sensor saturation. 

\begin{figure}[!t]
  \centering
  \includegraphics[width=\columnwidth]{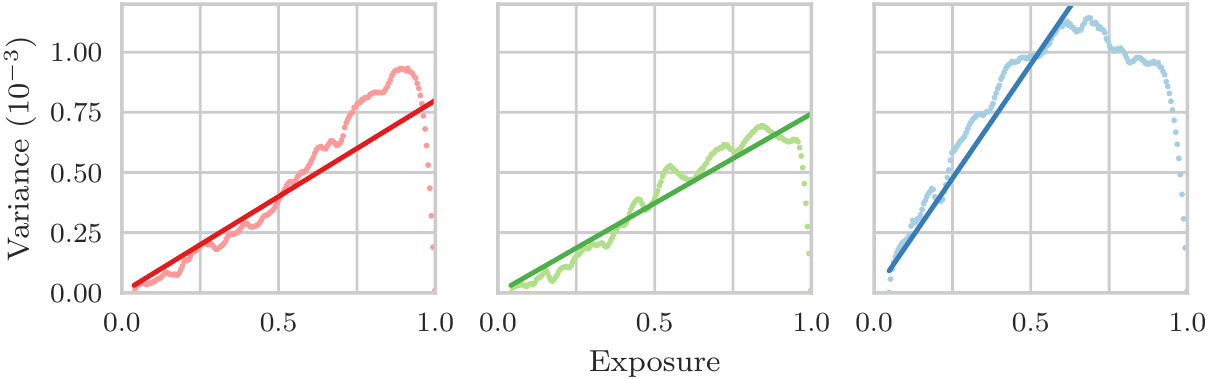}
  \caption
  {
    Dependency between exposure and its variance for different color channels. Straight lines through the origin are fitted to the data points to evince the suitability of the noise model \eref{imaging-equation}.
  }
  \label{fig:exposure-variance}
\end{figure}

The slope of the fitted lines corresponds to the coefficient $a$ in \eref{radiance-uncertainty}. As expected, it is significantly larger for the blue channel. Our experiments show that it depends on the gain setting of the camera. Since our system operates with a fixed gain, the exact relation is not important.

\section{Incremental HDR color reconstruction}
\label{sec:inc}

In line with the incremental nature of SLAM, we formulate an online HDR color reconstruction scheme. As the mapping progresses, the camera delivers a stream of color images $Z_i$ taken with exposure times $t_i$. Consider a scene point $\mathbf{x}$ that projects onto the pixel locations $\mathbf{u}_i$ in these images. The task is to maintain an estimate of the point radiance and its uncertainty by merging the observations $z_i = Z_i(\mathbf{u}_i)$ as they become available.

Each pixel observation $z_i$ is an RGB triplet; we consider it \emph{valid} if all color channels are well-exposed and \emph{invalid} otherwise. Invalid pixels are never fused into the radiance estimate, even if some color channels are within the saturation limits. Effectively, this means that updates to all color channels of a point radiance estimate are synchronized. This is in contrast with the standard HDR imaging, where color channels are treated in isolation.

This rule is motivated by the fact that the camera motion and imprecisions in tracking cause errors in pixel-to-point association. Thus, independent updates of color channels may introduce arbitrary distortion of the apparent point color. In a batch processing system this problem can be addressed by consistency tests \cite{Granados2013}. In the incremental setting this can not be resolved entirely; synchronized updates reduce the impact allowing only blur-like distortions.

An immediate consequence of this rule is that the color of a reconstructed scene point can be in one of the two states: \emph{incomplete} and \emph{complete}. The color is created incomplete; it turns complete when the first valid observation of this point is made. A toy example of such evolution is given in \fref{evolution}. Below we formally define the performed operations based on the state of the color and observation.

\begin{figure*}[!t]
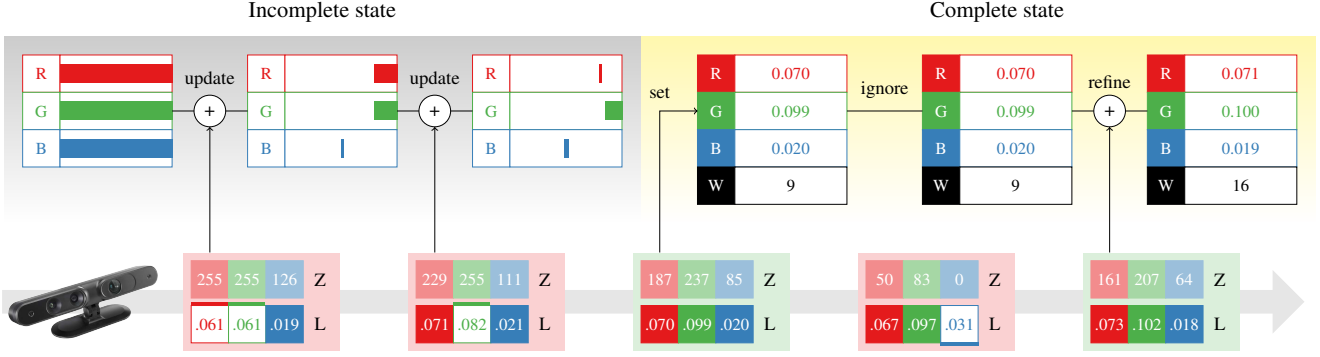

  \centering
  \includestandalone[width=\textwidth]{evolution}
  \vspace*{3pt}
  \caption
  {
    A HDR color (top row) evolves over time as new pixel observations (bottom row) become available. Upon creation, the color is incomplete and no bounds on its radiance are known (indicated by full color bars). The arriving invalid observations (indicated by red background) are used to update the radiance bounds. For example, the overexposed red channel of the first observation means that the radiance of the red channel is at least 0.061. When the first valid observation becomes available (indicated by green background), the color turns complete and the bounds are replaced with exact radiance estimates and a confidence value. All subsequent invalid observations are ignored, whereas the valid observations are averaged in, increasing the certainty of the radiance estimate.
  }
  \label{fig:evolution}
\end{figure*}

\subsection{Incomplete state}

Upon creation and until the first valid observation becomes available, a point is in the incomplete state, meaning that it has no exact radiance estimate. Instead, it is represented by a tuple of ranges \tuple{\lambda^{R},\lambda^{G},\lambda^{B}} that bound the radiance of each color channel and inform the decisions of the exposure time controller (as detailed in \cref{sec:ctrl}).

Starting from the whole effective dynamic range of the camera $\lambda_{\text{sys}} = \left[l^{\text{min}},l^{\text{max}}\right]$ the upper and lower bounds are progressively refined using the information contained in the invalid observations. An important insight is that underexposed channels give an upper bound on the radiance, while overexposed channels give a lower bound.

Below we define how the bounds are computed from invalid pixels. All color channels are treated the same way, thus we only discuss a single channel and drop the superscripts to simplify the notation. Suppose that a set of observations $\mathcal{Z}=\{z_i\}$ was made and a set $\mathcal{L}$ of their radiances was computed using \eref{radiance}. Let $\check{\mathcal{L}}$, $\hat{\mathcal{L}}$, and $\bar{\mathcal{L}}$ be subsets containing radiances from under-, over-, and well-exposed observations respectively. From these data, the radiance range
\begin{equation}
  \lambda =
  \begin{cases}
    \left[\max(\hat{\mathcal{L}} \cup l^{\text{min}}),\min(\check{\mathcal{L}} \cup l^{\text{max}})\right] & \text{if }\bar{\mathcal{L}}=\varnothing\\
    \left[\min(\bar{\mathcal{L}}),\max(\bar{\mathcal{L}})\right] & \text{otherwise}.
  \end{cases}
  \label{eqn:incomplete}
\end{equation}
Note that this is trivially translated into incremental updates.

\subsection{Complete state}

After the first valid observation becomes available, a point is in the complete state, meaning that it has an exact radiance estimate. In order to reduce the variance of this estimate, new valid observations are averaged in.

Under the assumption of compound Gaussian noise, Granados \etal[Granados2010] have shown that optimal HDR reconstruction is achieved if observations are weighted using the inverse of sample variance. In their system the photon and dark current shot noises are modeled, leading to a circular dependency between the estimates of radiance and sample variance. Thus an iterative optimization is required, ruling out online operation.

Differently from them, in our simplified camera model the dark current shot noise is excluded and the sample variance is assumed to be proportional to the true radiance. Substituting the radiance \eref{radiance} and inverse of sample variance from \eref{radiance-uncertainty} as weight into \eref{hdr}, we obtain the following formula for the radiance estimate after the $k^{\text{th}}$ measurement:
\begin{equation}
  \bar{L}_k =
    \frac{\displaystyle\sum_{i}^{k}\frac{t_{i}v_{i}}{aL}\frac{g(z_{i})}{t_{i}v_{i}}} {\displaystyle\sum_{i}^{k}\frac{t_{i}v_{i}}{aL}} =
    \frac{\displaystyle\sum_{i=1}^{k}g(z_{i})}{\displaystyle\sum_{i=1}^{k}t_iv_i},
  \label{eqn:complete-radiance}
\end{equation}
where $v_i = V(\mathbf{u}_i)$ are the vignetting attenuation factors at the locations of pixel observations.  This estimate can be updated incrementally as new valid observations become available \cite{West1979}. The value in the denominator is the accumulated weight $w_k$ of the radiance estimate.

\section{Map-aware exposure time control}
\label{sec:ctrl}

The long-term goal in HDR mapping is to obtain a reliable color estimate for each reconstructed scene point.
In this section we describe a controller that chooses an exposure time with maximum expected utility in the next frame.

There are two types of points in the map: incomplete and complete. Ideally, each point should become complete and have high SNR. Therefore, the objective for incomplete points is to obtain a valid observation, and for complete points is to increase their accumulated weight.

The camera motion is not controlled by the mapping system; it has no knowledge of the planned trajectory. The only reasonable assumption is that the motion is locally smooth and the velocity is such that two consecutive frames capture almost the same part of the scene. This allows to restrict the control decisions to be based only on a subset of the map visible in the last frame. Conveniently, the view prediction component is already a part of the SLAM loop; it renders the current state of the reconstruction into the image space $\Omega\subset\mathbb{N}^{2}$ for the purposes of frame-to-model tracking. Below we assume that besides from the depth map, it produces another three maps: radiance $L:\Omega\rightarrow\mathbb{R}^3$, weight $W:\Omega\rightarrow\mathbb{R}$, and radiance bounds $\Lambda:\Omega\rightarrow\mathbb{R}^6$.

We define an utility function that evaluates the expected gain of choosing a particular exposure time $t$ given the rendered state of the reconstruction: 
\begin{equation}
  U(t,L,W,\Lambda)=U_{e}(t,\Lambda)+U_{r}(t,L,W).
  \label{eqn:utility}
\end{equation}
It consists of two terms, exploration $U_e$ and refinement $U_r$. The former is targeted at the incomplete points and analyzes the radiance bounds map $\Lambda$; the latter is concerned with the complete points and analyzes the radiance and weight maps $L$ and $W$. The balance between the exploration and refinement can be adjusted by scaling one of the terms.

\subsection{Exploration}

The controller aims to turn incomplete points into complete by finding an exposure time that allows to get a valid observation. This search is guided by the estimates of boundaries on radiance maintained for each incomplete point, as described in the previous section.

We assume that the true radiance of a point is log-uniformly distributed within the radiance bounds $\lambda$. Therefore, given the detectable range $\lambda_t$ of a certain exposure time $t$, the probability that the point will be observed without saturation can be computed as 
\begin{equation}
  p(\lambda,\lambda_{t}) =
    \frac{\left\langle\lambda\cap\lambda_{t}\right\rangle}{\left\langle\lambda\right\rangle},
\end{equation}
where $\left\langle\cdot\right\rangle$ denotes the interval length in log-space. Since each point has three color channels, a product of the individual channel probabilities has to be computed. Denoting the subset of pixel locations of incomplete points as $\mathcal{I}$, the exploration utility is therefore defined as:
\begin{equation}
  U_{e}(t,\Lambda) =
    \sum_{\mathbf{u}\in\mathcal{I}}\prod_{c\in\left\{ R,G,B\right\} }p(\Lambda^{c}(\mathbf{u}),\lambda_{t}^{c}).
\end{equation}

\subsection{Refinement}

A complete point has an exact, albeit noisy, estimate of its radiance. It can be improved by integrating additional, preferably low-variance, samples. The goal is, thus, for each point to get a valid observation at maximum possible exposure time, as it will have highest possible weight.

As discussed above, our attention is limited to the points visible in the previous frame. Their radiances and accumulated weights were rendered into the $L$ and $W$ maps. Assuming that the points will project to approximately the same pixel locations in the next frame, and since vignetting effects expose spatially smooth variations, we expect to receive irradiance $E = LV$ at the sensor. Depending on the exposure time, some of these will fall in the detectable range and give valid observations. For exposure time $t$, let
\begin{equation}
  \mathcal{V}_{t}=\left\{\mathbf{u}\in\Omega\mid E(\mathbf{u})\in\epsilon_t\right\}
  \label{eqn:valid-pixels}
\end{equation}
be a subset of pixels that will have valid observations. They will be fused into the model. The contribution of each observation is proportional to exposure time and inversely proportional to the weight already accumulated by the point. Thus, we define the refinement utility as:
\begin{equation}
  U_r(t,L,W)=\sum_{\mathbf{u}\in\mathcal{V}_{t}}\frac{t}{W(\mathbf{u})}.
  \label{eqn:refine}
\end{equation}

\section{Implementation details}
\label{sec:impl}

We based our system on the open-source implementation of ElasticFusion~\cite{Whelan2015}. It is not HDR-aware and works with colors in LDR image space, conventionally representing them as 24-bit RGB triplets. However, the space allocated for each color is 64 bits. By fitting our HDR color representation into this space, we avoid any impact on the memory footprint. The complete colors are represented by 3 radiances and a common weight, thus 16-bit integers are used, which is sufficient to represent the full dynamic range supported by the system. The incomplete colors are represented by a zero weight and 3 radiance ranges (\ie 6 numbers), each truncated into 8-bit integers.

How quickly a camera reacts to the changes in exposure time setting (\ie control lag) is of high practical importance. The \Asus cameras that we have tested respond to the control commands within 3 frames. Therefore, while tracking and data fusion run at the full framerate, the controller is limited to approximately 10 Hz.

The base SLAM implementation utilizes dense direct odometry with geometric and photometric residuals as a tracking front-end. In our implementation the photometric residuals is lifted into the HDR color space.

\section{Experimental evaluation}
\label{sec:eval}

\subsection{Exposure time selection with a static camera}
\label{sub:exposure-time-selection-on-static-scene}

We quantitatively demonstrate the efficiency of the proposed exposure time controller in comparison with a set of baselines. The baseline controllers sweep through the allowed exposure time range in upward and downward direction with either multiplicative or additive steps.

We fix the camera in front of a high dynamic range scene and perform HDR reconstruction using the standard batch-processing approach to obtain the ground truth (see \fref{bunny}). Next we perform incremental HDR reconstruction using the method described in \cref{sec:inc} and selecting next exposure time with our controller and a set of baselines. After fusing each frame the mean reconstruction error \wrt the ground truth and the fraction of complete points is recorded.

\begin{figure}[!t]
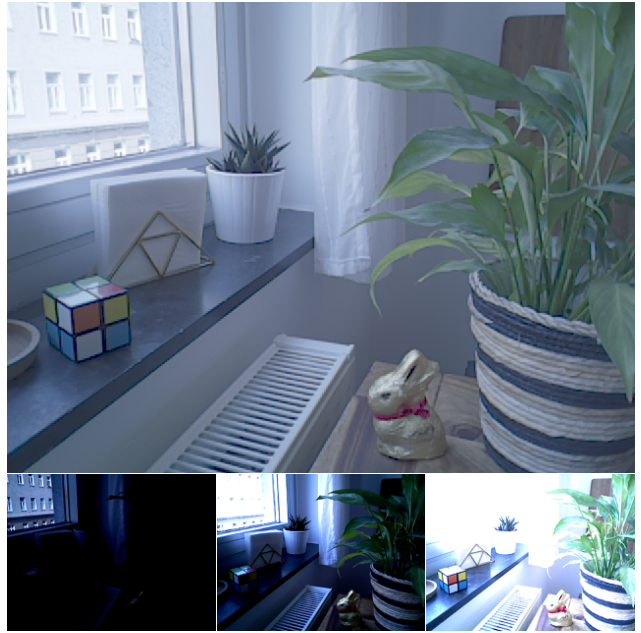

  \centering
  \includestandalone[width=\columnwidth]{bunny}
  \caption
  {
    Top: ground truth reconstruction of a high dynamic range scene. Bottom row: subset of used images (taken with minimum, mid-range, and maximum exposure times).
  }
  \label{fig:bunny}
\end{figure}

\fref{controller-evaluation} demonstrates the obtained results. Our controller explores the scene faster, leaving no incomplete points after observing 4 frames. The mean reconstruction error also decreases faster, reaching a steady state of about 2\% after integrating 15 frames.

\begin{figure}[!t]
  \centerline
  {
    \includegraphics[width=1.0\columnwidth]{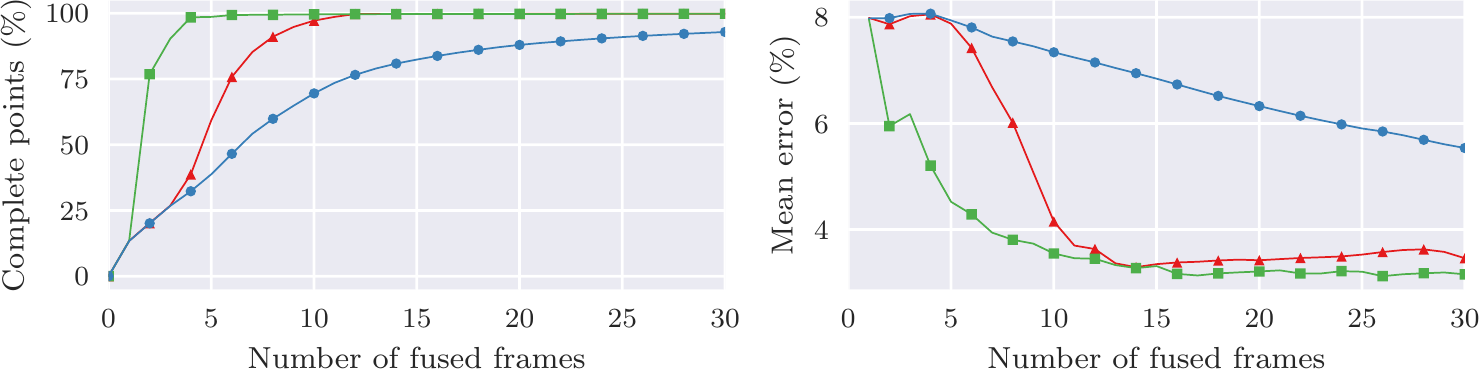}
  }
  \caption
  {
    Evaluation of the controllers in terms of the percentage of complete points and mean reconstruction error. Our controller (\ColorGreen[$\blacksquare$]), multiplicative controller (\ColorRed[$\blacktriangle$]), and incremental controller (\ColorBlue[$\bullet$]).
  }
  \label{fig:controller-evaluation}
\end{figure}

\subsection{HDR reconstruction with a moving camera}

We qualitatively demonstrate the performance of our system by reconstructing several office scenes and comparing the results with the maps produced by vanilla ElasticFusion with and without AEC. \fref{cmp-entrance,cmp} present side-by-side comparisons. The LDR reconstructions have numerous artifacts in their color textures. With disabled AEC (\fref{cmp-entrance}), the insufficiency of dynamic range of the LDR colors is manifested in overexposed white surfaces that appear to have the same color in the reconstruction. With enabled AEC (\fref{cmp}), the changes in exposure time are not accounted for by the LDR system and manifest in both strong and smooth color gradients on the walls. The HDR reconstructions do not have such defects.

\begin{figure}[!t]
  \centering
  \subfloat{\includegraphics[width=\columnwidth]{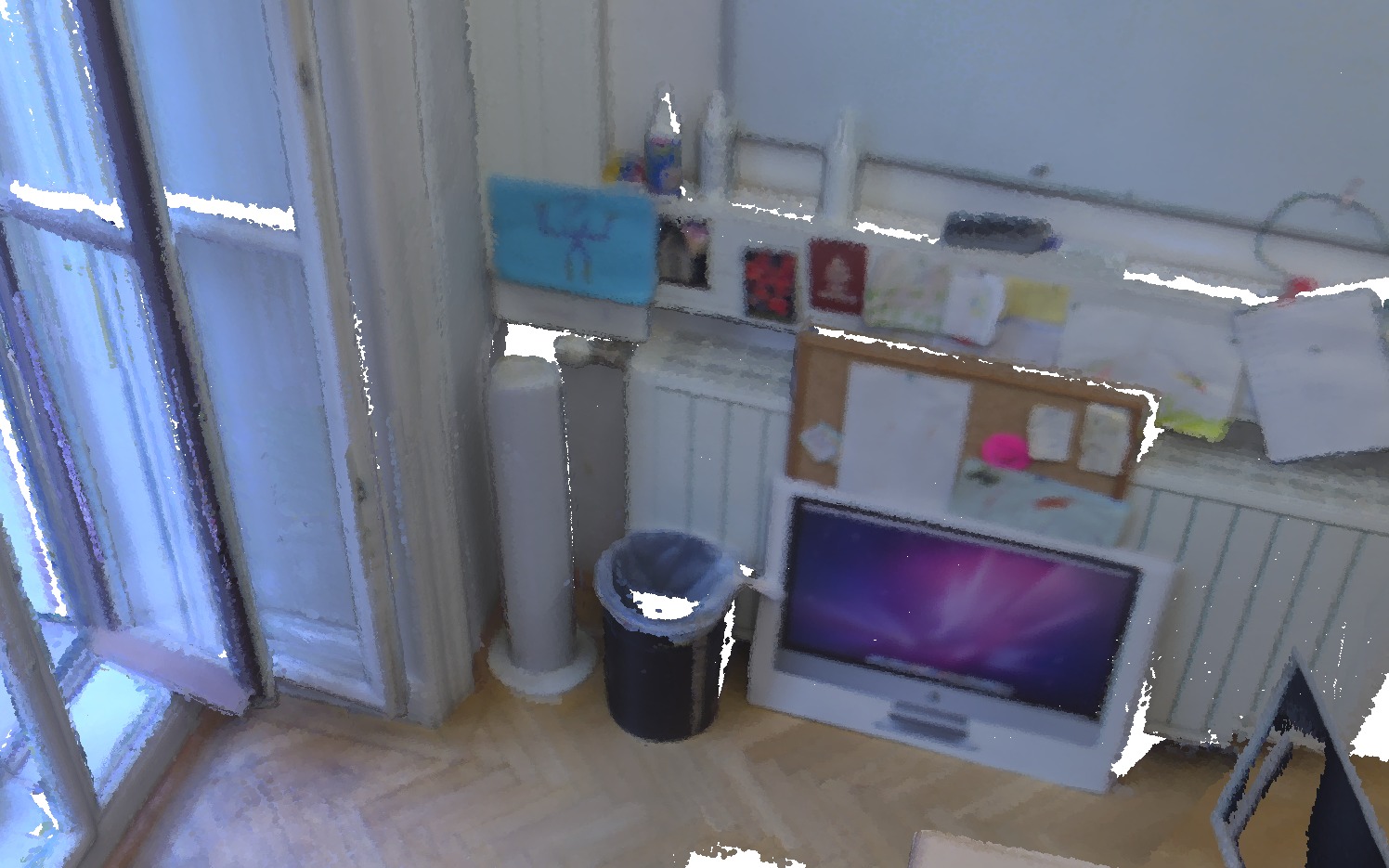}}%
  \vspace*{-6pt}
  \subfloat{\includegraphics[width=\columnwidth]{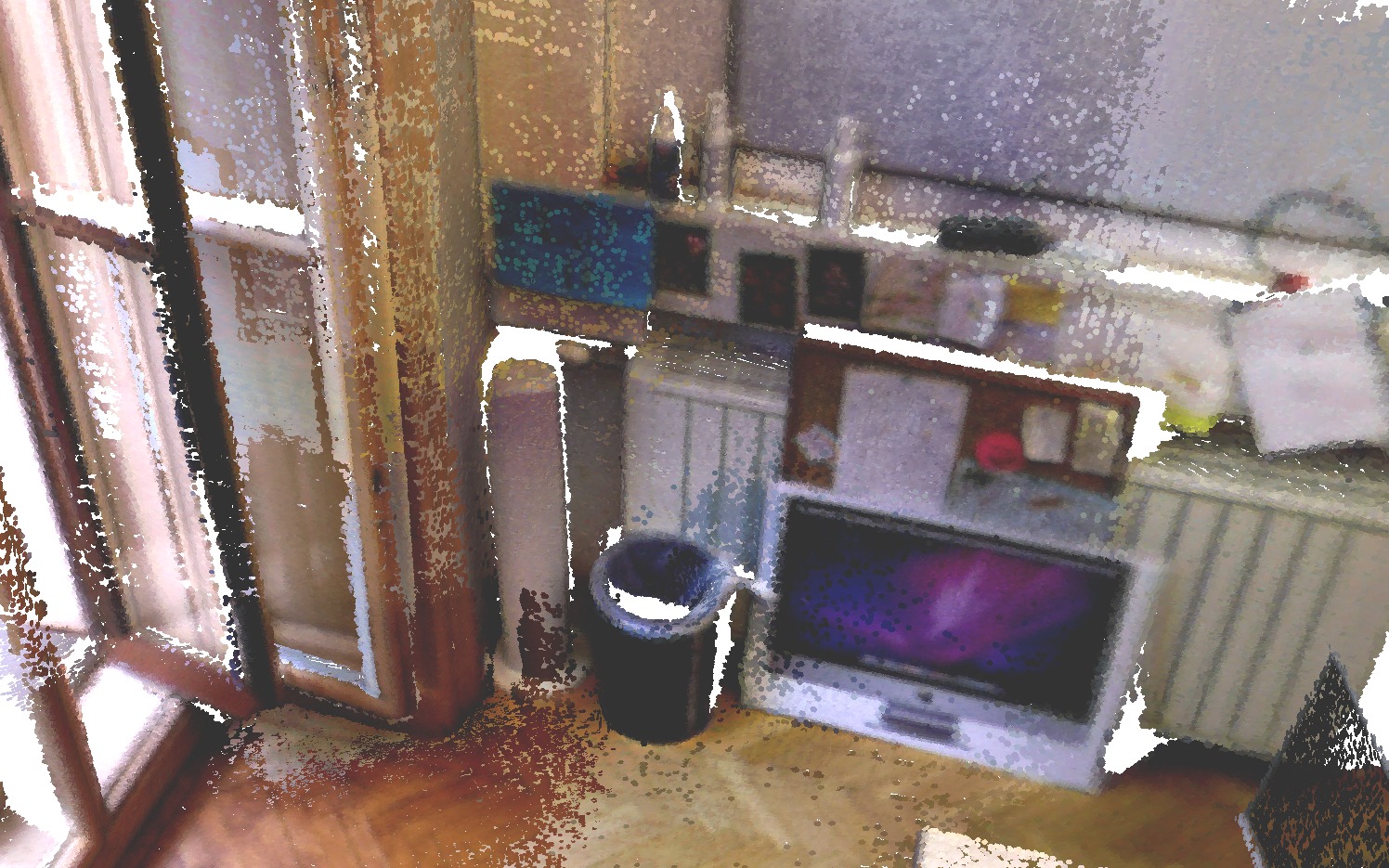}}%
  \vspace*{-6pt}
  \subfloat{\includegraphics[width=\columnwidth]{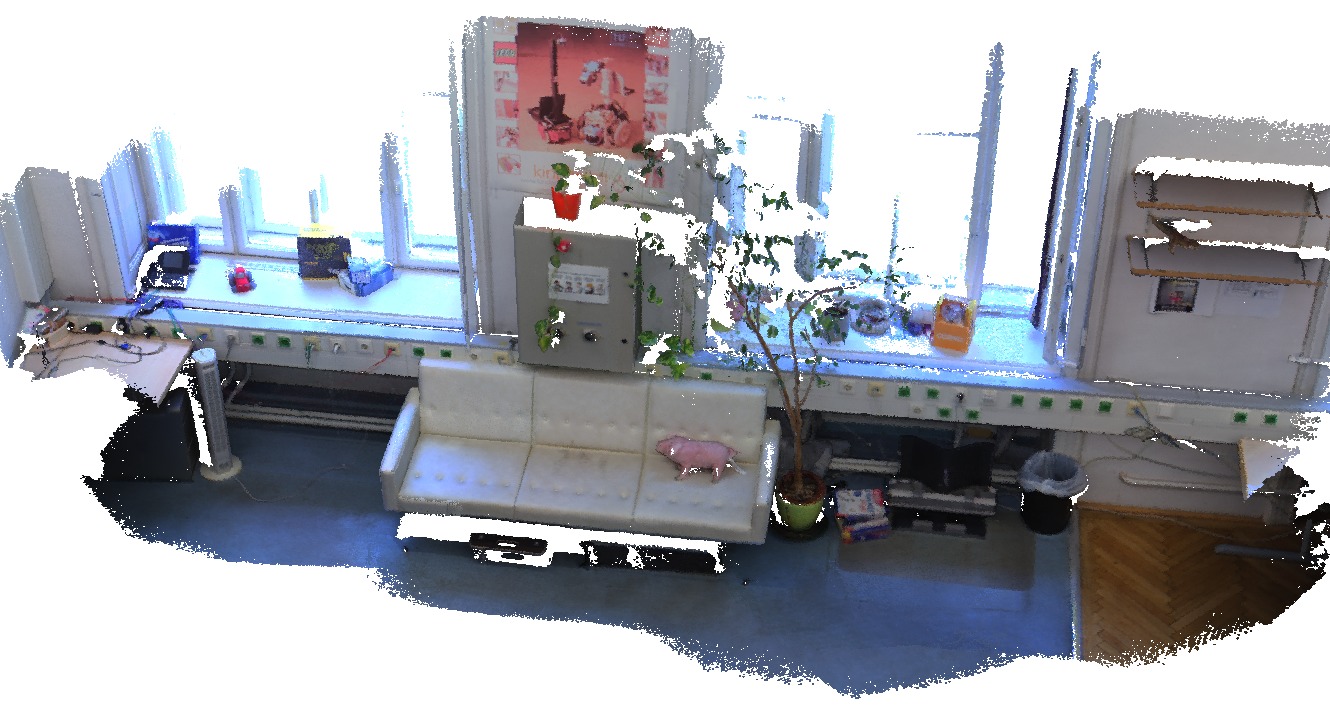}}%
  \vspace*{-6pt}
  \subfloat{\includegraphics[width=\columnwidth]{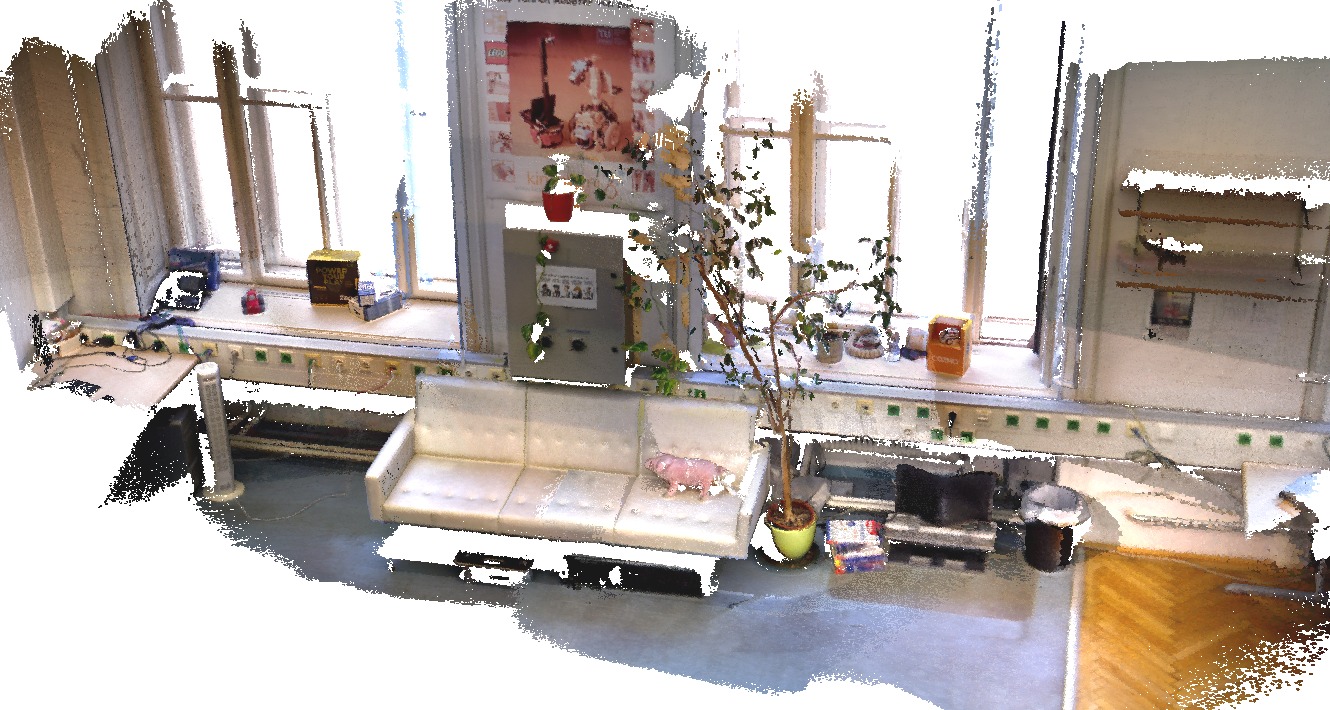}}%
  \caption{
    First and third: office scenes reconstructed in HDR with the proposed exposure time controller. Second and fourth: the same scenes reconstructed in LDR with ElasticFusion using camera built-in AEC function. Note the abrupt changes in color texture on the walls and on the floor in LDR reconstructions.
  }
  \label{fig:cmp}
\end{figure}

\section{Conclusions and future work}
\label{sec:ciao}

We presented an HDR-aware dense 3D reconstruction system. It leverages full radiometric camera calibration and relies on a simplified noise model tailored for the off-the-shelf \RGBD sensors. We introduced a concept of incomplete/complete colors that allows incremental HDR color fusion not common for classical HDR imaging methods. We also introduced an active exposure time controller into the mapping loop. It analyzes the reconstructed map to make decisions and maximize information gain in the next frames. In a set of experiments we demonstrated an improved visual quality of color appearance in acquired models compared to a baseline LDR system.

In the future it will be interesting to evaluate the impact that improved HDR textures have on the tracking performance. Another research direction would be  to address the changes in scene illumination and reflective materials.

\section*{Acknowledgments}

The work presented in this paper has been funded by the European Union Seventh Framework Programme (FP7/2007-2013) under grant agreement No. 600623 ("STRANDS") and No. 610532 ("SQUIRREL"). We thank the anonymous reviewers for their helpful comments.

\clearpage

{\small
\bibliographystyle{ieee}
\bibliography{references}
}

\end{document}